\documentclass[letterpaper, 10 pt, conference]{ieeeconf}  

\IEEEoverridecommandlockouts                              

\usepackage{times}
\usepackage{epsfig}
\usepackage{graphicx}
\usepackage{amsmath}
\usepackage{amssymb}

\usepackage{verbatim}
\usepackage[acronym]{glossaries}
\usepackage[super]{nth}
\usepackage{booktabs}
\usepackage{xcolor}
\usepackage{notations}

\usepackage[pagebackref=true,breaklinks=true,colorlinks,bookmarks=false]{hyperref}

\newcommand{\parsection}[1]{\vspace{2mm}\noindent\textbf{#1:}~}
\newcommand{\etc}{etc.}
\newcommand{\beginsupplementary}{%
        \setcounter{section}{0}
        \setcounter{table}{0}
        \renewcommand{\thetable}{S\arabic{table}}%
        \setcounter{figure}{0}
        \renewcommand{\thefigure}{S\arabic{figure}}%
     }

\title{\LARGE \bf
Local Memory Attention for Fast Video Semantic Segmentation
}

\author{Matthieu Paul, Martin Danelljan, Luc Van Gool and Radu Timofte$^{1}$
\thanks{$^{1}$All authors are with the Computer Vision Lab, ETH Z{\"u}rich, Switzerland. Corresponding author email: {\tt\footnotesize paulma@vision.ee.ethz.ch}}%
}

\begin{document}

\maketitle
\thispagestyle{empty}
\pagestyle{empty}

\begin{abstract}
   
We propose a novel neural network module that transforms an existing single-frame semantic segmentation model into a video semantic segmentation pipeline. In contrast to prior works, we strive towards a simple, fast, and general module that can be integrated into virtually any single-frame architecture. Our approach aggregates a rich representation of the semantic information in past frames into a memory module. Information stored in the memory is then accessed through an attention mechanism. In contrast to previous memory-based approaches, we propose a fast local attention layer, providing temporal appearance cues in the local region of prior frames. We further fuse these cues with an encoding of the current frame through a second attention-based module. The segmentation decoder processes the fused representation to predict the final semantic segmentation. We integrate our approach into two popular semantic segmentation networks: ERFNet and PSPNet. We observe an improvement in segmentation performance on Cityscapes by 1.7\% and 2.1\% in mIoU respectively, while increasing inference time of ERFNet by only 1.5ms. Source code is available at \url{https://github.com/mattpfr/lmanet}.

\end{abstract}

\section{Introduction}

\begin{figure}[t]
\centering%
\includegraphics[width=0.78\linewidth]{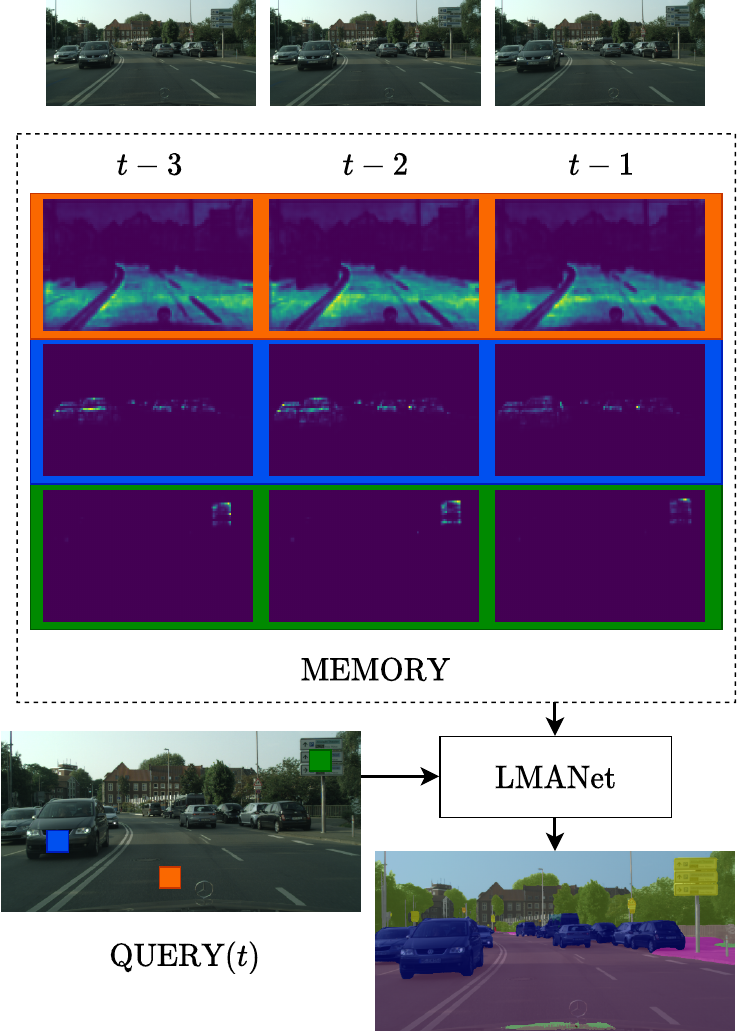}\vspace{-3mm}
\caption{We propose LMANet for fast video semantic segmentation. The Memory (top) consists of deep features extracted from previous frames. Features of the current frame (Query) are matched against the frames in the Memory to generate memory attention maps for each query location. Three example locations (orange, blue, green) are visualized with their respective attention maps in Memory. Our approach can thus efficiently integrate information from past frames to predict a final segmentation output.}%
\label{fig:marketing}\vspace{-4mm}
\end{figure}

Semantic segmentation is one of the core computer vision tasks that paves the way towards scene understanding. It entails assigning a label to each pixel of an image, where a label generally represents an element in a predefined set of \emph{classes}. It is useful in a growing number of applications, including augmented reality, surveillance and robotics (autonomous cars and drones), where scene understanding is key to build better and safer systems. In these scenarios, however, the input to the vision pipeline most often consists of a video stream, and not only a single frame. It is natural to exploit the temporal dimension instead of processing each input frame independently. In this work, we therefore address the problem of \emph{video} semantic segmentation.

Considering a sequential input instead of a single-frame can be beneficial to solve the semantic segmentation task, as we can leverage the additional temporal dimension. Indeed, in most cases, one can reasonably assume a certain level of temporal consistency from one frame to the next. This temporal cue has the potential to reduce inference time by avoiding redundant computations, improve the accuracy of the prediction, or find a better trade-off between speed and accuracy. In practice however, exploiting video data is a very challenging problem. Introducing the temporal dimension can easily lead to inertia, which can be problematic for dynamic content. In particular, the model should avoid propagating errors over time and allow for rapid changes in the scene produced by events such as occlusions or new objects. Efficiency is also a concern with video architectures, as they need to run fast enough and require a lot of data to train and evaluate.

In comparison to its single-frame counterpart, video semantic segmentation has received much less attention despite its potential advantages. Existing approaches exploit video input by using either dense optical flow directly~\cite{Xu_2018_CVPR, Zhu2017DeepFF, Paul_2020_WACV, gadde2017semantic} or by combining it with \glspl{rnn}~\cite{tokmakov:hal-01511145, Nilsson_2018_CVPR}. However, these often lead to marginal improvements relative to the added complexity and increased frame rates of those architectures. In contrast, methods trying to reduce complexity and inference time by avoiding redundancies and propagating information between frames usually have unsatisfactory accuracy compared to their baselines~\cite{Paul_2020_WACV, Li_2018_CVPR, Xu_2018_CVPR}. In this work, we set out to address these issues.

We introduce a novel neural network architecture to convert single-frame semantic segmentation models into video pipelines. 
Our network consists of three components. Following the recent success of attention-based approaches~\cite{Oh_2019_ICCV,li2019attention,10.1007/978-3-030-58607-2_21,hu2020temporally, Hu2021RealTimeSS}, we employ a \emph{Memory module} that aggregates semantic information from several past frames. It holds a rich representation of the past and is updated over time with incoming frames. Contrary to previous attention-based alternatives~\cite{li2019attention,hu2020temporally, Hu2021RealTimeSS}, we propose a local attention mechanism. It ensures efficient reading from the memory, given the current Query frame, while preserving the segmentation performance of the standard dot-product attention~\cite{NIPS2017_7181}.
We combine the extracted memory representation with features form the Query frame itself with an attention-based \emph{Fusion module}. The fused representation serves as input for the decoder to make the final prediction. Our approach does not rely on optical flow or other expensive operations, and thus has a minimal impact on inference time. We integrate our approach into two popular semantic segmentation architectures, namely ERFNet~\cite{Romera2018ERFNetER} and PSPNet~\cite{zhao2017pspnet}. With ERFNet, our approach achieves 1.67\% increase in mIoU on Cityscapes~\cite{Cordts2016Cityscapes} while only increasing inference time by 1.5ms.

\section{Related work}

The simplest way of tackling video semantic segmentation is to perform semantic segmentation independently in each frame. Although this approach ignores the temporal dimension of the input, it provides a natural baseline to assess the performance of video segmentation methods.
%
%
Dedicated video semantic segmentation pipelines have been receiving more attention in the recent years. However, generating accurate dense annotations for video semantic segmentation is costly. Only synthetic data sets such as GTA5~\cite{Richter_2016_ECCV} or Sintel~\cite{Butler:ECCV:2012} provide large-scale densely  annotated training and test sets. Thus, most real-world video data sets are sparsely annotated~\cite{Brostow2009SemanticOC, Geiger2013IJRR, Cordts2016Cityscapes}.
Consequently, most video semantic segmentation approaches~\cite{Mahasseni2017BudgetAwareDS,10.1007/978-3-319-49409-8_69, Li_2018_CVPR, Xu_2018_CVPR, Paul_2020_WACV, Zhu2017DeepFF, tokmakov:hal-01511145, Nilsson_2018_CVPR, gadde2017semantic} evaluate their results based on a single frame annotated per sequence. Their basis is a single-frame classifier, such as a random forest or a CNN. Temporal consistency is introduced by considering consecutive predictions or propagating semantic information. The goal is to either improve the segmentation masks or to decrease the average inference time.

Some methods address video semantic segmentation by looking at the problem from a broader angle: using tracking ~\cite{Lezama2011}, motion segmentation ~\cite{Ochs:2014:SMO:2693343.2693376}, or even exploiting the 3D structure of the scene. The latter usually rely on 3D point clouds to improve 2D segmentation predictions ~\cite{10.1007/978-3-540-88682-2_5, 6248007, Kundu2014JointSS, brostow2008segmentation} which is computationally expensive and potentially error-prone. Other methods focusing on improving the segmented outputs model the temporal consistency by introducing inter-frame and intra-frame pixel connections in large graphs structures, mostly relying on 2D or 3D conditional random fields across the video inputs~\cite{5711561, Nijs2012OnlineSP, 7780714, Tripathi2015SemanticVS}.

Propagating semantic information and features was explored in different ways, for instance by using \glspl{lstm}~\cite{Mahasseni2017BudgetAwareDS} to embed the temporal aspect in the pipeline structure or scheduling key frames to fully segment and propagate semantic information in between~\cite{10.1007/978-3-319-49409-8_69, Li_2018_CVPR, Xu_2018_CVPR, Paul_2020_WACV}. These methods in general offer different operating points leading to different trade-offs between efficiency and accuracy.

One of the popular ways to propagate information between frames is to explicitly use dense optical flow. This is usually done with two different objectives in mind. First it may be used to reduce the average inference time. The authors of~\cite{Xu_2018_CVPR, Zhu2017DeepFF} use state-of-the-art GPU methods~\cite{DFIB15, IMKDB17}, adding to the complexity of their pipeline. In~\cite{Paul_2020_WACV}, the authors suggest a simpler approach relying on a fast optical flow running on CPU~\cite{kroegerECCV2016} to propagate semantics across frames and achieve higher frame rates. However, all of these methods reduce the segmentation accuracy.
The second objective of using optical flow is to improve the baseline accuracy. For instance, some pipelines learn temporal consistency between consecutive feature maps with the help of \glspl{gru}~\cite{tokmakov:hal-01511145, Nilsson_2018_CVPR}. Others warp features directly at different depths of their baseline~\cite{gadde2017semantic}. Unfortunately, these frameworks have relatively small gains over their baselines, and the added complex and slow optical flow structure around them makes them unpractical to use in real-time scenarios.


The transformer model~\cite{NIPS2017_7181} was introduced as an alternative to \glspl{rnn}, such as \glspl{lstm} and \glspl{gru}. Transformer networks and self-attention are mostly used in the field of Natural Language Processing but are rapidly gaining interest in computer vision topics such as image recognition~\cite{hu2019local}, object detection~\cite{10.1007/978-3-030-58452-8_13}, video object segmentation~\cite{Oh_2019_ICCV} or semantic segmentation~\cite{10.1007/978-3-030-58607-2_21}. 
Attention mechanisms were recently used for video semantic segmentation to leverage temporal information and improve both intra-frame and inter-frame semantic information. The authors of~\cite{li2019attention} introduce two attention mechanisms to combine features while others embed the attention mechanism in the backbone over several frames~\cite{hu2020temporally, Hu2021RealTimeSS}. In contrast, we aim at a simpler architecture and integrate a local attention mechanism to ensure efficiency.

\section{Proposed Method}
\label{sec:proposed_method}

\begin{figure*}[t]
\centering
\includegraphics[width=1.0\linewidth]{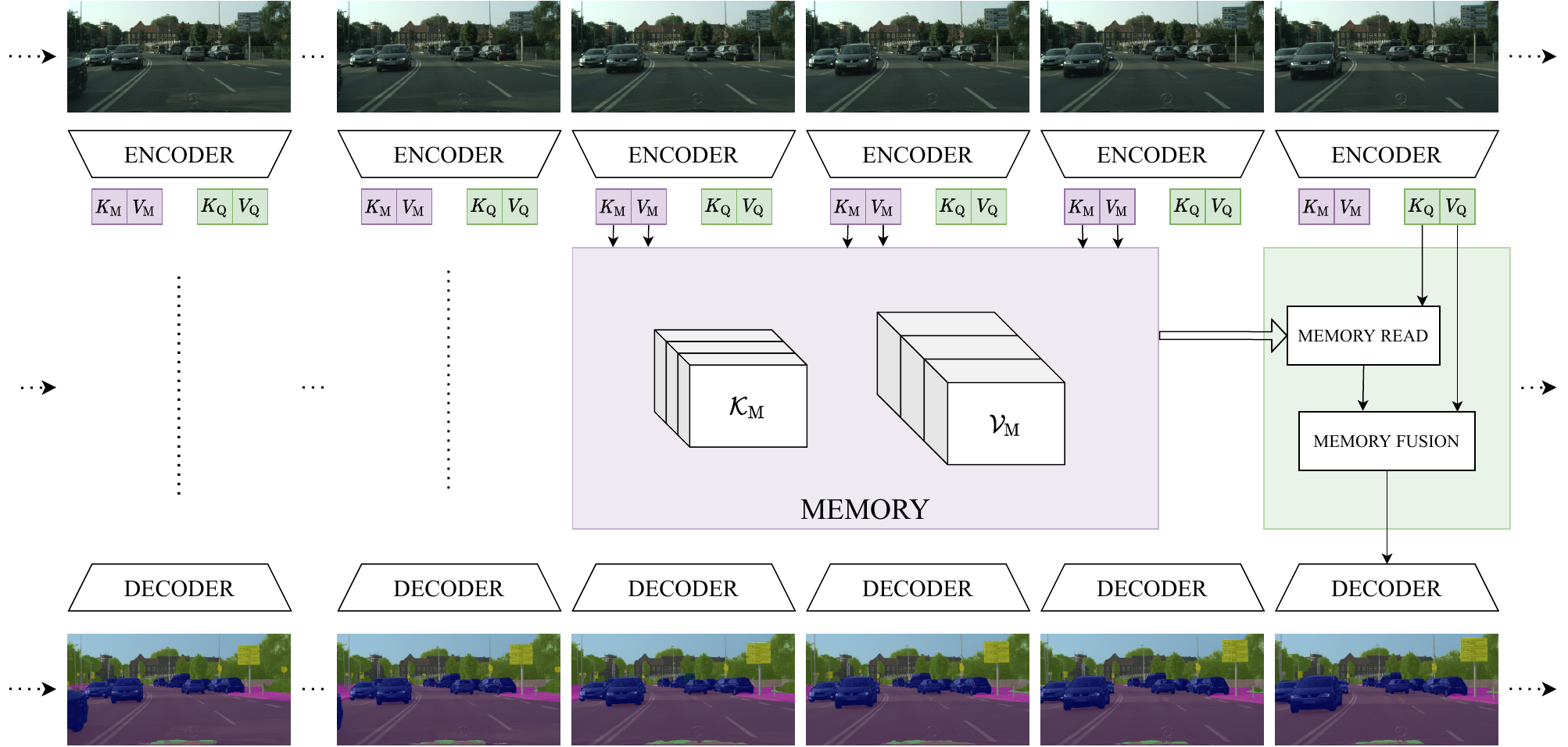}\vspace{-3mm}
\caption{Overview of our framework, built upon an existing Encoder-Decoder backbone. For each input frame, the Encoder output is used to create two \emph{(Key, Value)} pairs. One is stored in the \emph{Memory} module (violet) for future use and the other is used by the Memory Read and Fusion modules (green). The Memory Read module first accesses the relevant semantic information in the Memory. The information read from the memory is then combined with features of the current frame. The segmentation Decoder employs this fused representation to finally predict the segmentation mask.}%
\label{fig:overview}\vspace{-3mm}
\end{figure*}

We propose Local Memory Attention Networks (LMANet), that transforms an existing single frame semantic segmentation model into a video semantic segmentation pipeline. 
Our approach, summarized in Fig.~\ref{fig:overview}, aggregates an encoded representation from several past frames into a Memory module described in section \ref{subsec_method_mem_struct}. We describe how this memory is accessed via an attention-based module in section \ref{subsec_method_matching} and \ref{subsec_method_mem_read}. Finally, in section \ref{subsec_method_mem_fusion}, we detail how the features map from the Query frame are fused with the aggregated features read from the Memory to provide an input for the Decoder.

\subsection{Memory}
\label{subsec_method_mem_struct}

Having a representation of the temporal information from previous frames is a way to leverage the additional dimension offered by videos. Previous work trying to model this use a combination of optical flow and \glspl{gru}~\cite{tokmakov:hal-01511145, Nilsson_2018_CVPR}. However, \glspl{rnn} are often difficult to train, requiring long sequences. More importantly, the hidden state can easily get corrupted over time, severely affecting the performance during inference. In contrast, we propose to use a memory which holds semantic information from past frames as a core component of our framework. Following the success of attention based models for video semantic segmentation~\cite{li2019attention, hu2020temporally, Hu2021RealTimeSS}, we construct the memory using \gls{keys} and \gls{values} for each cell, such that the dimensionality of the Keys is much lower than the one of the Values. This strategy allows us to efficiently read out semantic information, stored in the values, by matching keys.  

We build and update the memory using consecutive inputs. For each input frame $I_t$, we generate 2 pairs of (Key,Value) features: one for the Memory ($\KM^t$, $\VM^t$) and one for the Query ($\KQ^t$, $\VQ^t$). Those features are all obtained from the same backbone Encoder output, each with its own simple convolution layer that preserves their spatial size, while reducing the dimensionality of the keys.
To keep a simple and explicit temporal representation of the Memory, the two sets of feature maps \gls{keys} and \gls{values} are obtained by concatenating the keys and values from the previous frames, as follows We define $\V = [\VM^1,\ldots,\VM^{L-1}] \in \reals^{\HM \times \WM \times \DV \times (L-1) }$ and $\K = [\KM^1,\ldots,\KM^{L-1}] \in \reals^{\HM \times \WM \times \DK \times (L-1) }$. With $\MEMIDXS = \{1,\ldots,L-1\}$ we denote the list of indices of the frames included in memory.

\subsection{Local Key Matching}
\label{subsec_method_matching}

In order to segment a new frame, we need to access the semantic information from the Memory. This is done by matching the feature maps from the Memory and the Query, respectively $\KM$ and $\KQ$. Those feature maps have a spatial size of $\HM \times \WM$ and a dimensionality of $\DK$. We let $\KM(i, j)$ and $\KQ(k, l)$ denote the features vectors of dimension $\reals^{\DK}$ at their respective spatial locations $(i,j), (k,l) \in \{1,\ldots,\HM\}\times\{1,\ldots,\WM\}$.

Previous approaches using transformers network architectures for NLP~\cite{NIPS2017_7181} or video object segmentation~\cite{Oh_2019_ICCV} perform an all-vs-all matching of their feature maps. However, this is memory-consuming and computationally expensive. The similarities between all pairs of spatial locations in the Memory and Query feature maps can be written as a 4D tensor $\CG(\KM, \KQ) \in \reals^{\HM \times \WM \times \HM \times \WM}$ defined as
\begin{equation}
\label{eq:global-corr}
\CG(\KM, \KQ) = \KM(i, j)^\text{T}\KQ(k, l).
\end{equation}
Particularly in our case, a global matching can be problematic for three reasons. The spatial size of feature maps and their dimensionality not only makes it harder to fit reasonable batch sizes and consecutive frames in GPU memory during training, but it also degrades inference time. Moreover, matching features at completely different locations can introduce wrong correlations between similar classes or instances in the scene. 

To address these issues, we exploit the temporal prior given by our input video. In general, the content of the current frame at a given position is more likely to be found in a similar position in the previous few frames. For each feature location in the current frame, we thus only read from the memory in its spatial neighborhood. We show that this operation can be efficiently implemented as a correlation layer, commonly used in optical flow networks~\cite{DFIB15}.

\begin{figure}[t!]
\centering
\includegraphics[width=\linewidth]{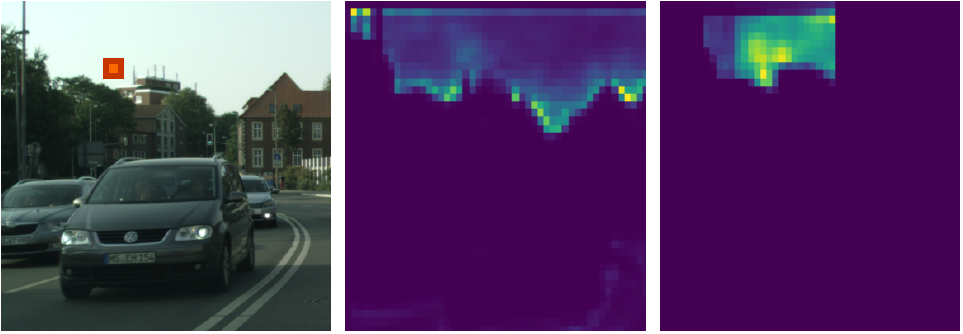}\vspace{-3mm}
\caption{For a given location (orange) in the query image (left), we compare the global correlation (middle) and the local correlation (right).}%
\label{fig:corrs-viz}\vspace{-4mm}
\end{figure}

Formally, the pairwise similarities between all the locations in the Query and their corresponding neighborhood in the Memory can be written as a 4D tensor $\CL(\KM, \KQ) \in \reals^{\HM \times \WM \times R \times R}$, where $R$ defines the radius of the local neighborhood from a given location. This can be expressed as
\begin{equation}
\label{eq:local-corr}
\CL(\KM, \KQ) = \KM(i, j)^\text{T} \KQ(i+k,j+l).
\end{equation}
In this case, $(i,j) \in \{1,\ldots,\HM\}\times\{1,\ldots,\WM\}$ and $(k,l) \in \REGION = \{-R,\ldots,R\}^2$ represents the search region for the correlation. $(k,l)$ therefore represents the displacement relative to the reference frame location $(i,j)$, constrained to a radius $R$. 

We choose $R$ such that $R << \HM,\WM$. While the local correlation cannot match features that are located outside the limited search region, we can immensely reduce the complexity in memory and computation time for feature maps of a large spatial size, from  $\mathcal{O}((\HM\WM)^2)$ to $\mathcal{O}((\HM\WM)\times R^2)$.

\subsection{Local Memory Attention}
\label{subsec_method_mem_read}

The Memory reading operation is the core part of our method. It enables us to access the relevant features contained in the Memory at the right location in an efficient way, thanks to the matching described in the previous section. It is performed in three steps. First, the Query key is matched against all the keys in the Memory. We thus define
\begin{equation}
\label{eq:mem-read-1}
\quad C^t_{kl}(i,j) = \CL(\KM^t, \KQ) = \KM^t(i, j)^\text{T} \KQ(i+k,j+l).
\end{equation}
Then, we transform feature correlation maps into a set of probability maps using a SoftMax layer, expressed as
\begin{equation}
\label{eq:mem-read-2}
\quad P^t_{kl}(i,j) = \frac{\exp(C^t_{kl}(i,j))}{\sum_{klt}\exp(C^t_{kl}(i,j))}.
\end{equation}
The resulting probability maps represent the multiplicative attention weights, visualized in Figure~\ref{fig:corrs-viz}. 
These weights are crucial for the last step of the Memory Reading operation as they allow us to weight the features read from the Memory in both spatial and temporal dimensions. The module thus only reads from the parts of the Memory that contain the most relevant information for each location. Formally, the output tensor $\MEMREAD$ read from the Memory at each location $(i,j)$ can be written as
\begin{equation}
\label{eq:mem-read-3}
\MEMREAD(i,j) = \sum_{t \in \MEMIDXS} \sum_{(k,l) \in \REGION} P^t_{kl}(i,j) \VM^t(i+k,j+l).
\end{equation}
The read operation is a weighted sum of all values of the memory $\VM^t$ over the temporal domain $\MEMIDXS$ and the spatial domain $\REGION$. By construction, its dimensions are identical to the dimensions of the Query value tensor: $\VQ,\MEMREAD \in \reals^{\HM \times \WM \times \DV}$.

\subsection{Memory Fusion}
\label{subsec_method_mem_fusion}

As described in the previous section, the semantic features read from the Memory $\MEMREAD$ contain a rich information from the aggregation of the semantics of previous frames. The features extracted in the Query $\VQ$ contain the encoded semantic segmentation of the current frame. The role of the Fusion module is to combine $\VQ$ and $\MEMREAD$ and provide a meaningful input the Decoder that can then make the most accurate prediction for the current frame. 
The naive approach that would consist in concatenating feature maps would not allow the network to select the best features from both sources of information. Indeed, the model ideally has to learn to trust more the Memory for parts of the scene that have a consistent behaviour over time (static classes, regular or slow motions, \etc) while taking into account the current frame features that might introduce sudden changes (objects entering the scene, occlusions, irregular motions, \etc).

\begin{figure}[t]
\centering
\includegraphics[width=\linewidth]{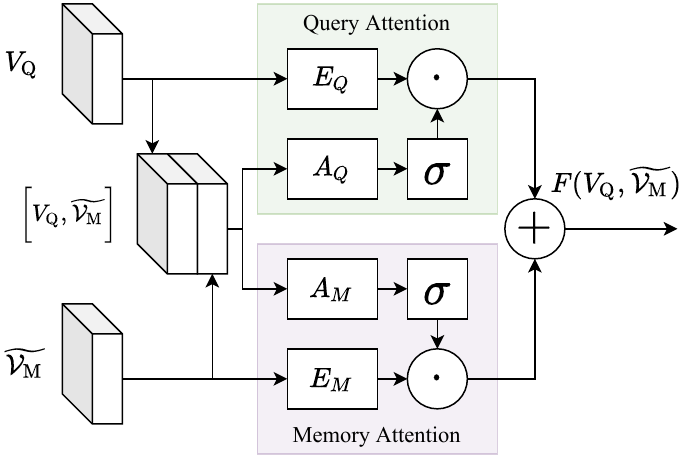}\vspace{-3mm}
\caption{Fusion of the features read from the memory and the current frame.}
\label{fig:fusion}\vspace{-3mm}
\end{figure}

In order to overcome this limitation, we extend the concatenation solution by adding one attention branch per Fusion input, see Fig.~\ref{fig:fusion}. Each branch uses its corresponding Fusion input and the concatenated inputs. Those attention branches, with the help of sigmoid activated gates inspired by RNNs, are controlling the information flowing through them. This structure therefore allows to select which features from the Query or from the Memory will be sent to the Decoder. Formally, if we let $\sigma$ be the sigmoid activation function and '$\cdot$' be the element-wise tensor multiplication, the output of the Fusion module can be written as
\begin{align}
\label{eq:mem-fusion}
F(\VQ, \MEMREAD) &= \sigma\big(\AQ(\VQ, \MEMREAD)\big) \cdot \EQ(\VQ) \nonumber \\
                           &+ \sigma\big(\AM(\VQ, \MEMREAD)\big) \cdot \EM(\MEMREAD).
\end{align}
The Fusion output $F(\VQ, \MEMREAD) \in \reals^{\HM \times \WM \times \DENC}$ is a 3D tensor with the same dimensions as the Encoder output. $\EQ, \AQ, \AM, \EM$ represent respectively the learned convolution layers of the Query attention and Memory attention branches, as shown in Figure~\ref{fig:fusion}.

\subsection{Training method}

Our goal is to enable single-frame semantic segmentation models to deal with video inputs, thus we need a simple and efficient training routine. However, two obstacles stand in the way. First, we have a limited amount of annotated data. Second, we need to fit a large enough amount of images on the GPU while keeping reasonable batch sizes.

Assuming that the backbone Encoder and Decoder are trained beforehand on single frames, we propose to train our module in two steps. First, we freeze the parameters of both the Encoder and the Decoder while we train our Memory layers and Fusion module from scratch for a small amount of epochs. We choose the number of epochs between 20 and 50, so that the accuracy of our model reaches the range of the baseline accuracy without aiming for convergence. This allows the layers of our modules to get a rough initialization before we move on to the next step, where we jointly train our modules and the Decoder until convergence.

\section{Evaluation}
\label{sec:evaluation}

\subsection{Experimental setup}
\label{ssc:setup}

We run all our experiments on the Cityscapes dataset~\cite{Cordts2016Cityscapes}, as it is commonly used to benchmark both single frame and video semantic segmentation methods. It contains diverse urban street scenes across 50 different cities which makes it relevant for autonomous driving applications. Its training, validation and testing sets contain respectively 2975, 500 and 1525 images with a resolution of $1024\times2048$ pixels. Each of those images represent the \nth{20} frame of a corresponding 30 frames video sequence. The training and validation sets provide dense ground truth annotations, but only for the \nth{20} frame of each sequence. We therefore train and evaluate our approach on video snippets but report mean Intersection over Union (mIoU) results on the provided \nth{20} frame.

Our experimental setup consists of several Intel Xeon Gold 6242 CPUs running at 3.5GHz and several NVIDIA GeForce RTX 2080Ti GPUs. We use Python 3.7, PyTorch 1.2, CUDA 10.2 and cuDNN 7.6. While simple inference can be run on a single GPU, training is usually performed on several GPUs to either speed up training or overcome GPU memory limitations. For training, we use the Adam optimizer with an initial learning rate of 2e-4. We use an adaptive scheduling based on the train loss: the learning rate is kept constant until the loss reaches a minimum plateau, after which we decrease it by 20\%. We keep this strategy until the learning rate reaches a minimum value of 1e-8.

\subsection{Ablation study}
\label{ssc:ablation}

We perform an ablation study to highlight the role of each component of our method and its impact on the performance of the pipeline. We thus focus on the critical parameters of our modules: the search region for the correlation layer, the size of the Memory used during inference and the fusion strategy. We run this ablation study using ERFNet~\cite{Romera2018ERFNetER} as backbone for efficiency reasons.

\parsection{Memory size $\MEMIDXS$} The first and foremost important aspect of our method is the size of its Memory. Especially in the context of autonomous driving, it is important to make sure to look at the past frames without holding on too old information. At the same time, we have to make sure that the memory is long enough to have meaningful temporal information to use. This part of the study aims at determining the impact of the memory size $S$ on the final prediction accuracy. The case $S=0$ represents our single-frame baseline ERFNet~\cite{Romera2018ERFNetER}. Table~\ref{tab:comp_mem_size} summarizes the results and highlights two aspects. First, all memory sizes improve the baseline from $+1.29\%$ to $+1.67\%$, which means that there is indeed meaningful information to retain from past frames and that our method uses it to improve the final prediction. Second, even though a single frame in the memory already improves the baseline, best results are achieved with a bigger memory, with a peak for $S=4$.

\begin{table}[h]\vspace{1mm}
\caption{Impact of the size of the Memory on the results. }
\vspace{-3mm}
\centering
\resizebox{1.0\columnwidth}{!}{%
\begin{tabular}{lcccccc}
\toprule
mIoU(\%)           & $S=0$ & $S=1$ & $S=2$ & $S=3$ & $S=4$ & $S=5$ \\ \midrule
LMA, $R=21$        & 72.05 & 73.34 & 73.41 & 73.37 & 73.72 & 73.46 \\ \bottomrule
\end{tabular}}
\label{tab:comp_mem_size}\vspace{-3mm}
\end{table}

\parsection{Search Region $\REGION$} This part of the study aims at validating the design choice of the keys matching step in our pipeline. As it is the pillar supporting the Memory reading operation, we want to compare an all-vs-all matching to a local one and measure the impact of the search region size on the resulting mIoU. Table~\ref{tab:comp_search_regions} summarizes the results. These numbers should be compared to the single frame baseline ($S=0$) which has a mIoU of $72.05\%$.

Here we investigate and validate three aspects of our approach. First, all settings combinations yield final mIoU improvements compared to the single frame baseline, ranging from $+0.85\%$ to $+1.67\%$. Second, we observe the importance of actually considering a local neighborhood for the features matching when considering temporal information: matching at the same location ($R=1$) improves results from the baseline, but is far from the improvements observed with bigger search regions. Third, we confirm that using a global matching is counter productive, as local matching with a search radius between $R = 21$ and $R = 41$ seems to yield the best results. This suggests that local matching with a reasonable search region $\REGION$ is the best option to avoid mismatches that could happen across the whole image while accounting for the temporal information accumulated in the Memory.

\begin{table}[h]\vspace{1mm}
\caption{Accuracy comparison between global ($G$) and local correlations with different search region sizes ($R<<G$).}\vspace{-3mm}
\centering
\resizebox{1.0\columnwidth}{!}{%
\begin{tabular}{lccccc}
\toprule
mIoU(\%)   & $\CG$   & $R=41$ & $R=21$ & $R=11$ & $R=1$ \\  \midrule
  $S=3$    & 73.38   & 73.50  & 73.37  & 73.21  & 72.90 \\
  $S=4$    & 73.49   & 73.64  & 73.72  & 73.21  & 72.94 \\ \bottomrule
\end{tabular}}
\label{tab:comp_search_regions}\vspace{-3mm}
\end{table}

\parsection{Fusion Strategy} The last part of our method combines features from the Query and the Memory, which hold different kinds of information. The former contains the current frame features while the latter is an aggregation of the information from past frames. Table~\ref{tab:comp_fusion_strategies} compares our results to a simple concatenation of the semantic maps and confirms that a more complex fusion with attention branches generates more accurate results ($+0.33\%$). This seems to indicate that the probabilistic weighting done by the Fusion is helpful to guide the Decoder.

\begin{table}[h]\vspace{1mm}
\caption{mIoU comparison between concatenation and our Fusion.}\vspace{-3mm}
\centering
\resizebox{0.9\columnwidth}{!}{%
\begin{tabular}{lccc}
\toprule
Baseline         & $\text{LMA}_{S=3, R=21}$ (Concat.) & $\text{LMA}_{S=3, R=21}$ (Fusion) \\  \midrule
72.05\%          & 73.14\%       &  73.37\%      \\ \bottomrule
\end{tabular}}
\label{tab:comp_fusion_strategies}\vspace{-3mm}
\end{table}

\subsection{Memory and computation time}

As we strive towards a simple and general method for fast video semantic segmentation applications, an important aspect to tackle is its efficiency, both in terms of GPU memory consumption and inference time. We therefore study the impact of our method on those variables on the existing backbone ERFNet~\cite{Romera2018ERFNetER}.  As reported previously, a relatively small memory size of 3 to 5 frames is reasonable for our benchmarks, so we focus on the impact of the search radius $R$, with a fixed memory size of $S=3$ and report the results in Table~\ref{tab:benchmark}.

\begin{table}[h]\vspace{1mm}
\caption{Inference time and GPU memory consumption on the 500 validation sequences of Cityscapes~\cite{Cordts2016Cityscapes}. Each sequence has a length of $L=4$ frames and the Memory size $S=3$. Results are displayed as an average of 10 runs under the same conditions.}\vspace{-3mm}
\centering
\resizebox{1.0\columnwidth}{!}{%
\begin{tabular}{lccccc}
\toprule
                    & Baseline & $\CG$    & $R=41$   & $R=21$   & $R=11$     \\  
\midrule
Total time T(s)      &  62.4    &  96.8     & 114      & 71.4       & 67.5       \\
$\Delta T$           &   -      &  + 55\%   & + 83\%    & + 14\%    & + 8\%     \\  
GPU Mem. M(MiB)      &  1553    &  4125     & 2116      & 1749      & 1606      \\
$\Delta M$           &   -      &  + 166\%  & + 36\%    & + 13\%    & + 3\%     \\ 
\bottomrule
\end{tabular}}
\label{tab:benchmark}\vspace{-3mm}
\end{table}

\begin{table*}[t]\vspace{1mm}
\caption{Comparison between our methods and other existing methods. As we focus on the temporal aspect, we report the single-frame baseline for each method and the relative changes it brings in terms of mean IoU and inference time. The fields marked with "n.a" denote data that was not available in the corresponding publications.}\vspace{-3mm}
\centering
\resizebox{1.0\linewidth}{!}{%
\begin{tabular}{l|lcc|ccccc}
\toprule
& \multicolumn{3}{c}{Single-frame baseline} & \multicolumn{5}{c}{Video approach} \\
Method                             & Backbone                           & mIoU (\%)  & Time (ms) & mIoU (\%)  & Time (ms) & $\Delta \text{mIoU}(\%)$ & $\Delta \text{T}(ms)$ & Resolution       \\
\midrule
$\text{LMA}_{S=4, R=21}$           & ERFNet~\cite{Romera2018ERFNetER}   &   72.05    &   10.1    &    73.72   &  11.6     & +1.67                    &  +1.5                 & $1024\times512$  \\
$\text{LMA}_{S=3, R=21}$           & PSP-SS-SC~\cite{zhao2017pspnet}    &   76.34    &   406     &    78.48   &   758     & +2.14                    &  +352                 & $2048\times1024$ \\
\midrule
TDNet~\cite{hu2020temporally}      &  BiseNet*18~\cite{hu2020temporally}&   73.8     &   20      &    75.0    &   21      &   +1.2                   &  +1                   & n.a.             \\
TDNet~\cite{hu2020temporally}      &  BiseNet*34~\cite{hu2020temporally}&   76.0     &   27      &    76.4    &   26      &   +0.4                   &  -1                   & n.a.             \\
FANet~\cite{Hu2021RealTimeSS}      &  FANet-18~\cite{Hu2021RealTimeSS}  &   75.0     &   14      &    75.5    &   14      &   +0.5                   &  +0                   & $2048\times1024$ \\
FANet~\cite{Hu2021RealTimeSS}      &  FANet-34~\cite{Hu2021RealTimeSS}  &   76.3     &   17      &    76.7    &   17      &   +0.4                   &  +0                   & $2048\times1024$ \\

GRFP~\cite{Nilsson_2018_CVPR}      & PSP-SS-MC~\cite{zhao2017pspnet}    &   79.7     &  n.a      &    80.2    &  n.a      &   +0.5                   &  +335                 & $512\times512$   \\
GRFP~\cite{Nilsson_2018_CVPR}      & PSP-MS-MC~\cite{zhao2017pspnet}    &   80.9     &  n.a      &    81.3    &  n.a      &   +0.4                   &  +335                 & $512\times512$   \\
NetWarp~\cite{gadde2017semantic}   & PSP-SS-MC~\cite{zhao2017pspnet}    &   79.4     &  3000     &    80.6    &  3040     &   +1.2                   &  +40                  & $713\times713$   \\
NetWarp~\cite{gadde2017semantic}   & PSP-MS-MC~\cite{zhao2017pspnet}    &   80.8     &  30300    &    81.5    &  30500    &   +0.7                   &  +200                 & $713\times713$   \\
LLVS~\cite{Li_2018_CVPR}           & ResNet-101~\cite{He2015}           &   80.2     &  360      &    76.84   &   171     &  -3.36                   &  -189                 & n.a              \\
DVS~\cite{Xu_2018_CVPR}            & PSP-SS-SC~\cite{zhao2017pspnet}    &   77.0     &  588      &    70.2    &    87     &  -6.8                    &  -501                 & n.a              \\
DVS~\cite{Xu_2018_CVPR}            & PSP-SS-SC~\cite{zhao2017pspnet}    &   77.0     &  588      &    62.6    &    33     &  -14.4                   &  -555                 & n.a              \\
DFF~\cite{Zhu2017DeepFF}           & ResNet-101~\cite{He2015}           &   71.1     &  658      &    69.2    &   179     &  -1.9                    &  -479                 & $1024\times512$  \\
EVS~\cite{Paul_2020_WACV}          & ICNet~\cite{Zhao_2018_ECCV}        &   67.3     &   26      &    66.2    &    13     &  -1.1                    &  -13                  & $2048\times1024$ \\
EVS~\cite{Paul_2020_WACV}          & ICNet~\cite{Zhao_2018_ECCV}        &   67.3     &   26      &    67.6    &    27     &  +0.3                    &  +1                   & $2048\times1024$ \\
\bottomrule
\end{tabular}}
\label{tab:comp_other_methods}\vspace{0mm}
\end{table*}

These measurements confirm that an all-vs-all matching of the feature maps is not a viable option for our problem. Indeed, a $166\%$ increase in GPU memory is not desirable during inference in a real-life scenario. It also would complicate the learning phase, where bigger batch sizes are usually beneficial. At the same time, we see that our pipeline adds a very small memory footprint onto the existing backbone when using a local correlation approach, only 13\% for $R=21$, which yields the highest mIoU.

Furthermore, we show that the search radius $R$ is crucial when it comes to inference efficiency, and that a bigger correlation region directly impacts the inference time, while not necessarily yielding the best accuracy results. For a search radius of 21 pixels, we get a limited impact both on the GPU memory (+13\%) and on the inference time (+14\%).

\begin{figure*}[t]
\centering
\includegraphics*[trim=0 430 0 0,width=\linewidth]{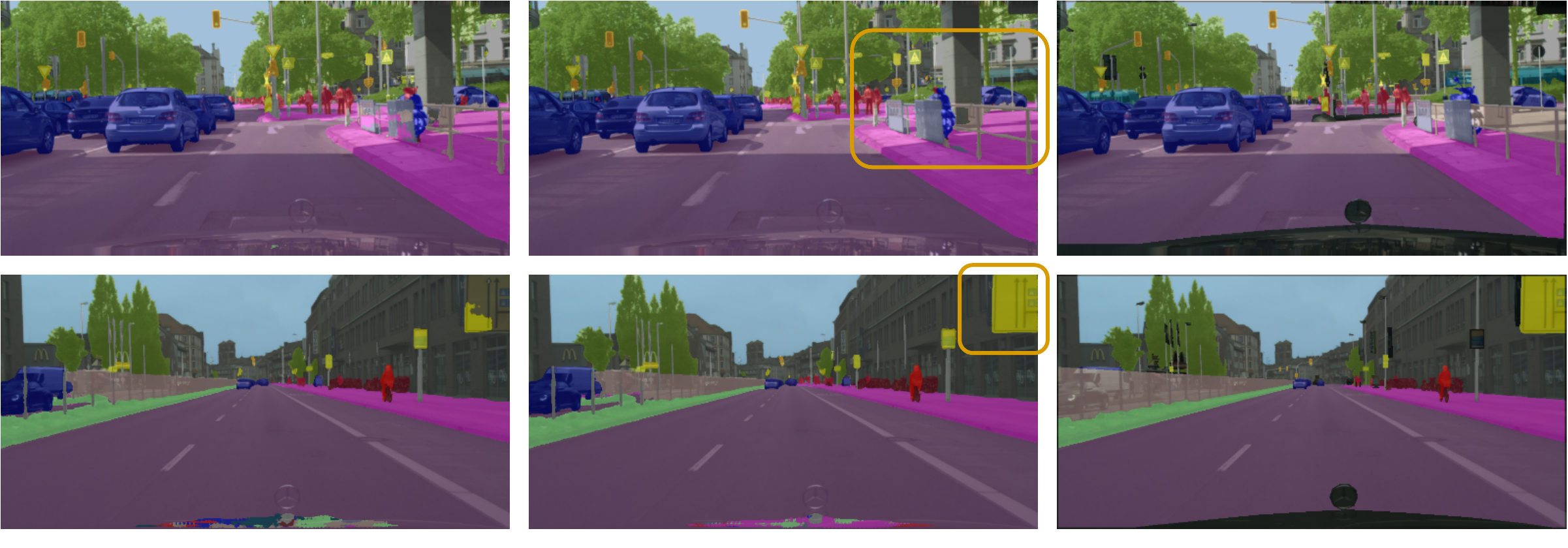}
\includegraphics*[trim=0 0 0 810,width=\linewidth]{figures/LMANet-Qualitative.pdf}\vspace{-3mm}
\caption{Qualitative results comparison between our method $\text{LMA}_{4,21}$ and the single-frame  ERFNet~\cite{Romera2018ERFNetER}. Different scenarios are highlighted in the orange boxes and show improvements over the baseline. The top example shows cleaner segmentation masks around a mixture of \textit{barrier}, \textit{motorbike}  and \textit{car}. The bottom example shows better classification of a \textit{road sign} at the edge of the scene.}%
\label{fig:qualitative}\vspace{0mm}
\end{figure*}

\begin{table*}[h]\vspace{1mm}
\caption{Comparison with two backbones (ERFNet and PSPNet) on class-wise improvements brought with our method LMA (Ours).}\vspace{-3mm}
\centering
\resizebox{1.0\linewidth}{!}{%
\begin{tabular}{l|c|ccccccccccccccccccc}
\toprule
Method  & Total & road & s.walk & build. & wall & fence & pole & tlight & sign & veg. & terrain & sky & person & rider & car & truck & bus & train & motorcycle & bicycle  \\  
\midrule
\midrule
ERFNet                 & 72.05	& 97.65	& 81.28	& 90.68	 &49.31	&54.89	&59.93	&62.31	  & 72.07 & 91.25  & 60.83  & 93.36	&75.91	 & 53.28 & 92.84  & 72.77   & 78.77	 & 63.78  & 46.35      & 71.74   \\
\midrule
+ LMA, $R=21,S=4$	   & 73.72	& 97.73	& 81.81	& 90.83	 &51.65	&55.32	&60.87	&63.48	  & 74.21 & 91.50  & 61.17  & 93.43	&77.08	 & 57.97 & 93.21  & 75.63   & 81.66	 & 69.51  & 51.18      & 72.48   \\
\midrule
\midrule
PSPNet      &  76.34 &  98.05 & 84.71 &	92.47 &	54.66 &	60.71 &	64.05 &	71.22 &	79.09 &	92.64 &	63.49 &	94.71 &	82.46 &	62.43 &	94.96 &	73.87 &	82.51 &	58.70 &	61.70 &	78.04	\\
\midrule
+ LMA, $R=21,S=3$	&  78.48 &  98.06 &	85.06 &	92.73 &	60.13 &	62.57 &	64.20 &	70.22 &	78.20 &	92.40 &	64.17 &	94.35 &	81.68 &	63.16 &	95.14 &	80.77 &	87.93 &	76.86 &	67.24 &	76.21   \\
\bottomrule
\end{tabular}}
\label{tab:classes_search_radius}
\end{table*}

\subsection{Comparison to state-of-the-art}

Since current state-of-the-art approaches conduct experiments  with different baselines, backbones, input resolutions and hardware, providing a completely fair comparison is virtually impossible. In order to provide the fairest comparison, we report results from the original paper of the single-frame baseline along with the corresponding video-based improvements.

Based on the ablation study, we choose the settings that yield the best mIoU, which is with a search radius $R=21$, and we compare our framework with other video semantic segmentation methods. To provide a better comparison and validate our approach on a different backbone, we apply our method to two of the most popular backbones, ERFNet~\cite{Romera2018ERFNetER} and PSPNet~\cite{zhao2017pspnet} and report results in Table~\ref{tab:comp_other_methods}.

\parsection{ERFNet} While it does not reach the current state-of-the-art semantic segmentation accuracy, ERFNet~\cite{Romera2018ERFNetER} remains a good compromise between frame rate and accuracy compared to heavier and slower methods such as PSPNet~\cite{zhao2017pspnet} and DeepLab~\cite{deeplabv3plus2018}. It is therefore important to see whether our approach can help increasing the accuracy of its predictions while keeping a low inference time. Table~\ref{tab:comp_other_methods} shows that our approach can improve the baseline mIoU by 1.67\% while increasing the runtime by only 1.5 ms. This gain in accuracy is bigger than any other existing methods and the very good inference time inherent to ERFNet~\cite{Romera2018ERFNetER} is preserved. In comparison, EVS~\cite{Paul_2020_WACV} improves its baseline by only 0.3\% with a similar additional inference time. Attention based approaches such as TDNet~\cite{hu2020temporally} or FANet~\cite{Hu2021RealTimeSS} bring relatively smaller mean IoU improvements to their backbones, at the price of more complex architectures. Other methods starting from baselines below 80\%~\cite{Xu_2018_CVPR, Zhu2017DeepFF} still yield a mIoU vs. inference time ratio that is much worse than our approach.

Figure~\ref{fig:qualitative} shows the qualitative improvements that our approach can bring through the usage of the Memory with attention mechanisms. In some situations, it increases the temporal consistency of the predictions for fixed objects, for instance \textit{barriers} and \textit{traffic lights}.

\parsection{PSPNet}The state-of-the-art and publicly available results for PSPNet~\cite{zhao2017pspnet} are achieved by running multiple inferences on image crops of resolution $713\times713$. In order to get the final prediction on the $1024\times2048$ resolution, the authors define a grid of 8 overlapping crops and combine the results of independent predictions. They also present results in two flavors, a single scale version that simply runs at the original scale and another that combines 6 different scales together. We refer in the table to the single-scale and multi-scale versions of the multi-crop inference as \emph{PSP-SS-MC} and \emph{PSP-MS-MC}, respectively.

As we propose a general framework to work with video inputs, such a multi-scale and multi-crop inference on each frame is not practical. First, from an architecture complexity point of view, the implications it would have on our query/memory structure would have to be studied further. Second, it would multiply the average inference time per frame. Therefore, we focus our study on a single-scale and single-crop version of PSPNet, as in DVSNet~\cite{Xu_2018_CVPR}. We refer it as \emph{PSP-SS-SC}. 
In this setup, our approach not only improves the accuracy of a complex backbone, but also generalizes to other single-frame models. Moreover, while our approach has a bigger impact on the inference time when using the PSPNet backbone, compared to the ERFNet case, the relative improvement we gain from our simple structure outperforms other more complex approaches.

\subsection{Class-wise analysis}

The mean intersection over union only offers a partial view on the improvements achieved by our approach. In this section, we propose an overview of the results we get with our LMANet for each class, on both backbones ERFNet~\cite{Romera2018ERFNetER} and PSPNet~\cite{zhao2017pspnet}, see Table~\ref{tab:classes_search_radius}.
Two aspects stand out with ERFNet~\cite{Romera2018ERFNetER}. First, the set of classes that were already well segmented (road, building, vegetation, terrain, sky, car) do not change too much from the baseline. They mostly get improved by a small margin, or slightly degrade. Second, smaller and dynamic classes that are the most critical to understand the dynamics of the scene get improvements ranging from 0.5\% to almost 6\%. This is especially relevant in the context of autonomous driving and flying.

With PSPNet~\cite{zhao2017pspnet}, the trend is similar but with some noticeable differences. While bigger classes, including moving dynamic ones (train, truck, bus, motorcycle, wall) are improving substantially, smaller objects do not benefit in the same manner as before. This might be due to the stronger feature representation provided by the ResNet-101 backbone along with the higher resolution of the feature map: $256\times128\times2048$ compared to $128\times256\times128$ for ERFNet.

\section{Conclusion}

We present a novel neural network architecture that transforms an existing single frame semantic segmentation model into a video semantic segmentation pipeline. In contrast to prior methods, we strive towards a simple structure that can easily be integrated and trained together with any single-frame semantic segmentation model. 
Our approach aggregates the semantic information from past frames into a memory module and uses local attention mechanisms both to access the features stored in the Memory and to fuse them with the input Query frame. We validate our approach on two popular semantic segmentation networks and show that we increase the segmentation performance of ERFNet on Cityscapes by almost 2\% while preserving its real-time performance.

\parsection{Acknowledgements}
This work was partly supported by uniqFEED AG, ETH Z\"urich Fund (OK), Huawei Technologies Oy (Finland) and Amazon AWS.


{\small
\bibliographystyle{ieee_fullname}
\bibliography{egbib}

\begin{thebibliography}{10}\itemsep=-1pt

\bibitem{Brostow2009SemanticOC}
Gabriel~J. Brostow, Julien Fauqueur, and Roberto Cipolla.
\newblock Semantic object classes in video: A high-definition ground truth
  database.
\newblock {\em Pattern Recognition Letters}, 30:88--97, 2009.

\bibitem{10.1007/978-3-540-88682-2_5}
Gabriel~J. Brostow, Jamie Shotton, Julien Fauqueur, and Roberto Cipolla.
\newblock Segmentation and recognition using structure from motion point
  clouds.
\newblock In David Forsyth, Philip Torr, and Andrew Zisserman, editors, {\em
  ECCV}, 2008.

\bibitem{brostow2008segmentation}
Gabriel~J Brostow, Jamie Shotton, Julien Fauqueur, and Roberto Cipolla.
\newblock Segmentation and recognition using structure from motion point
  clouds.
\newblock In {\em ECCV}, 2008.

\bibitem{Butler:ECCV:2012}
D.~J. Butler, J. Wulff, G.~B. Stanley, and M.~J. Black.
\newblock A naturalistic open source movie for optical flow evaluation.
\newblock In {\em ECCV}, 2012.

\bibitem{10.1007/978-3-030-58452-8_13}
Nicolas Carion, Francisco Massa, Gabriel Synnaeve, Nicolas Usunier, Alexander
  Kirillov, and Sergey Zagoruyko.
\newblock End-to-end object detection with transformers.
\newblock In Andrea Vedaldi, Horst Bischof, Thomas Brox, and Jan-Michael Frahm,
  editors, {\em ECCV}, 2020.

\bibitem{5711561}
A.~Y.~C. {Chen} and J.~J. {Corso}.
\newblock Temporally consistent multi-class video-object segmentation with the
  video graph-shifts algorithm.
\newblock In {\em WACV}, 2011.

\bibitem{deeplabv3plus2018}
Liang-Chieh Chen, Yukun Zhu, George Papandreou, Florian Schroff, and Hartwig
  Adam.
\newblock Encoder-decoder with atrous separable convolution for semantic image
  segmentation.
\newblock In {\em ECCV}, 2018.

\bibitem{Cordts2016Cityscapes}
Marius Cordts, Mohamed Omran, Sebastian Ramos, Timo Rehfeld, Markus Enzweiler,
  Rodrigo Benenson, Uwe Franke, Stefan Roth, and Bernt Schiele.
\newblock The cityscapes dataset for semantic urban scene understanding.
\newblock In {\em CVPR}, 2016.

\bibitem{DFIB15}
A. Dosovitskiy, P. Fischer, E. Ilg, P. H{\"a}usser, C. Haz{\i}rba{\c{s}}, V.
  Golkov, P. v.d. Smagt, D. Cremers, and T. Brox.
\newblock Flownet: Learning optical flow with convolutional networks.
\newblock In {\em ICCV}, 2015.

\bibitem{6248007}
G. {Floros} and B. {Leibe}.
\newblock Joint 2d-3d temporally consistent semantic segmentation of street
  scenes.
\newblock In {\em CVPR}, 2012.

\bibitem{gadde2017semantic}
Raghudeep Gadde, Varun Jampani, and Peter~V. Gehler.
\newblock Semantic video cnns through representation warping.
\newblock In {\em ICCV}, 2017.

\bibitem{Geiger2013IJRR}
Andreas Geiger, Philip Lenz, Christoph Stiller, and Raquel Urtasun.
\newblock Vision meets robotics: The kitti dataset.
\newblock {\em IJRR}, 2013.

\bibitem{He2015}
Kaiming He, Xiangyu Zhang, Shaoqing Ren, and Jian Sun.
\newblock Deep residual learning for image recognition.
\newblock In {\em CVPR}, 2016.

\bibitem{hu2019local}
Han Hu, Zheng Zhang, Zhenda Xie, and Stephen Lin.
\newblock Local relation networks for image recognition.
\newblock In {\em ICCV}, 2019.

\bibitem{hu2020temporally}
Ping Hu, Fabian Caba, Oliver Wang, Zhe Lin, Stan Sclaroff, and Federico
  Perazzi.
\newblock Temporally distributed networks for fast video semantic segmentation.
\newblock In {\em CVPR}, 2020.

\bibitem{Hu2021RealTimeSS}
P. Hu, Federico Perazzi, Fabian~Caba Heilbron, Oliver Wang, Zhe Lin, Kate
  Saenko, and S. Sclaroff.
\newblock Real-time semantic segmentation with fast attention.
\newblock {\em RA-L}, 6:263--270, 2021.

\bibitem{IMKDB17}
E. Ilg, N. Mayer, T. Saikia, M. Keuper, A. Dosovitskiy, and T. Brox.
\newblock Flownet 2.0: Evolution of optical flow estimation with deep networks.
\newblock In {\em CVPR}, 2017.

\bibitem{kroegerECCV2016}
Till Kroeger, Radu Timofte, Dengxin Dai, and Luc~Van Gool.
\newblock Fast optical flow using dense inverse search.
\newblock In {\em ECCV}, 2016.

\bibitem{Kundu2014JointSS}
Abhijit Kundu, Y. Li, F. Dellaert, F. Li, and James~M. Rehg.
\newblock Joint semantic segmentation and 3d reconstruction from monocular
  video.
\newblock In {\em ECCV}, 2014.

\bibitem{7780714}
A. {Kundu}, V. {Vineet}, and V. {Koltun}.
\newblock Feature space optimization for semantic video segmentation.
\newblock In {\em CVPR}, 2016.

\bibitem{Lezama2011}
J. Lezama, K. Alahari, J. Sivic, and I. Laptev.
\newblock Track to the future: Spatio-temporal video segmentation with
  long-range motion cues.
\newblock In {\em CVPR}, 2011.

\bibitem{li2019attention}
Jiangyun Li, Yikai Zhao, Jun Fu, Jiajia Wu, and Jing Liu.
\newblock Attention-guided network for semantic video segmentation.
\newblock {\em IEEE Access}, 7:140680--140689, 2019.

\bibitem{Li_2018_CVPR}
Yule Li, Jianping Shi, and Dahua Lin.
\newblock Low-latency video semantic segmentation.
\newblock In {\em CVPR}, 2018.

\bibitem{10.1007/978-3-030-58607-2_21}
Yifan Liu, Chunhua Shen, Changqian Yu, and Jingdong Wang.
\newblock Efficient semantic video segmentation with per-frame inference.
\newblock In Andrea Vedaldi, Horst Bischof, Thomas Brox, and Jan-Michael Frahm,
  editors, {\em ECCV}, 2020.

\bibitem{Mahasseni2017BudgetAwareDS}
Behrooz Mahasseni, Sinisa Todorovic, and Alan Fern.
\newblock Budget-aware deep semantic video segmentation.
\newblock {\em CVPR}, 2017.

\bibitem{Nijs2012OnlineSP}
R.~D. Nijs, Sebastian Ramos, Gemma Roig, X. Boix, L.~V. Gool, and K.
  K{\"u}hnlenz.
\newblock On-line semantic perception using uncertainty.
\newblock {\em IROS}, 2012.

\bibitem{Nilsson_2018_CVPR}
David Nilsson and Cristian Sminchisescu.
\newblock Semantic video segmentation by gated recurrent flow propagation.
\newblock In {\em CVPR}, 2018.

\bibitem{Ochs:2014:SMO:2693343.2693376}
Peter Ochs, Jitendra Malik, and Thomas Brox.
\newblock Segmentation of moving objects by long term video analysis.
\newblock {\em PAMI}, 36(6):1187--1200, June 2014.

\bibitem{Oh_2019_ICCV}
Seoung~Wug Oh, Joon-Young Lee, Ning Xu, and Seon~Joo Kim.
\newblock Video object segmentation using space-time memory networks.
\newblock In {\em ICCV}, 2019.

\bibitem{Paul_2020_WACV}
Matthieu Paul, Christoph Mayer, Luc~Van Gool, and Radu Timofte.
\newblock Efficient video semantic segmentation with labels propagation and
  refinement.
\newblock In {\em WACV}, 2020.

\bibitem{Richter_2016_ECCV}
Stephan~R. Richter, Vibhav Vineet, Stefan Roth, and Vladlen Koltun.
\newblock Playing for data: {G}round truth from computer games.
\newblock In {\em ECCV}, 2016.

\bibitem{Romera2018ERFNetER}
Eduardo Romera, J.~M. {\'A}lvarez, L.~M. Bergasa, and Roberto Arroyo.
\newblock Erfnet: Efficient residual factorized convnet for real-time semantic
  segmentation.
\newblock {\em T-ITS}, 19:263--272, 2018.

\bibitem{10.1007/978-3-319-49409-8_69}
Evan Shelhamer, Kate Rakelly, Judy Hoffman, and Trevor Darrell.
\newblock Clockwork convnets for video semantic segmentation.
\newblock In Gang Hua and Herv{\'e} J{\'e}gou, editors, {\em ECCV Workshops},
  2016.

\bibitem{tokmakov:hal-01511145}
Pavel Tokmakov, Karteek Alahari, and Cordelia Schmid.
\newblock {Learning Video Object Segmentation with Visual Memory}.
\newblock In {\em {ICCV}}, 2017.

\bibitem{Tripathi2015SemanticVS}
Subarna Tripathi, Serge~J. Belongie, Y. Hwang, and T. Nguyen.
\newblock Semantic video segmentation: Exploring inference efficiency.
\newblock {\em ISOCC}, pages 157--158, 2015.

\bibitem{NIPS2017_7181}
Ashish Vaswani, Noam Shazeer, Niki Parmar, Jakob Uszkoreit, Llion Jones,
  Aidan~N Gomez, \L~ukasz Kaiser, and Illia Polosukhin.
\newblock Attention is all you need.
\newblock In I. Guyon, U.~V. Luxburg, S. Bengio, H. Wallach, R. Fergus, S.
  Vishwanathan, and R. Garnett, editors, {\em Advances in Neural Information
  Processing Systems 30}, pages 5998--6008. Curran Associates, Inc., 2017.

\bibitem{Xu_2018_CVPR}
Yu-Syuan Xu, Tsu-Jui Fu, Hsuan-Kung Yang, and Chun-Yi Lee.
\newblock Dynamic video segmentation network.
\newblock In {\em CVPR}, 2018.

\bibitem{Zhao_2018_ECCV}
Hengshuang Zhao, Xiaojuan Qi, Xiaoyong Shen, Jianping Shi, and Jiaya Jia.
\newblock Icnet for real-time semantic segmentation on high-resolution images.
\newblock In {\em ECCV}, 2018.

\bibitem{zhao2017pspnet}
Hengshuang Zhao, Jianping Shi, Xiaojuan Qi, Xiaogang Wang, and Jiaya Jia.
\newblock Pyramid scene parsing network.
\newblock In {\em CVPR}, 2017.

\bibitem{Zhu2017DeepFF}
X. Zhu, Y. Xiong, Jifeng Dai, L. Yuan, and Y. Wei.
\newblock Deep feature flow for video recognition.
\newblock {\em CVPR}, 2017.

\end{thebibliography}
}

\clearpage
\section*{Supplementary Material}
\beginsupplementary
In this supplementary material, we provide detailed results and deeper analysis of the experiments performed in the paper. We provide the results obtained for each class and compare them for different operating points and backbones in section~\ref{part2-classe-wise-analysis}. Finally, we further investigate the memory size in section~\ref{part3-robustness}.

\section{Class-wise analysis}
\label{part2-classe-wise-analysis}

The mean intersection over union only offers a partial view on the improvements achieved by our approach. In this section we propose an overview of the results and improvements we get with our LMANet for each class, on both backbones ERFNet~\cite{Romera2018ERFNetER} and PSPNet~\cite{zhao2017pspnet} (Based on ResNet-101~\cite{He2015}). We also show the impact of the Memory size and search radius on those classes.

\subsection{ERFNet backbone}

\begin{figure*}[h]
\begin{center}
\includegraphics[width=\linewidth]{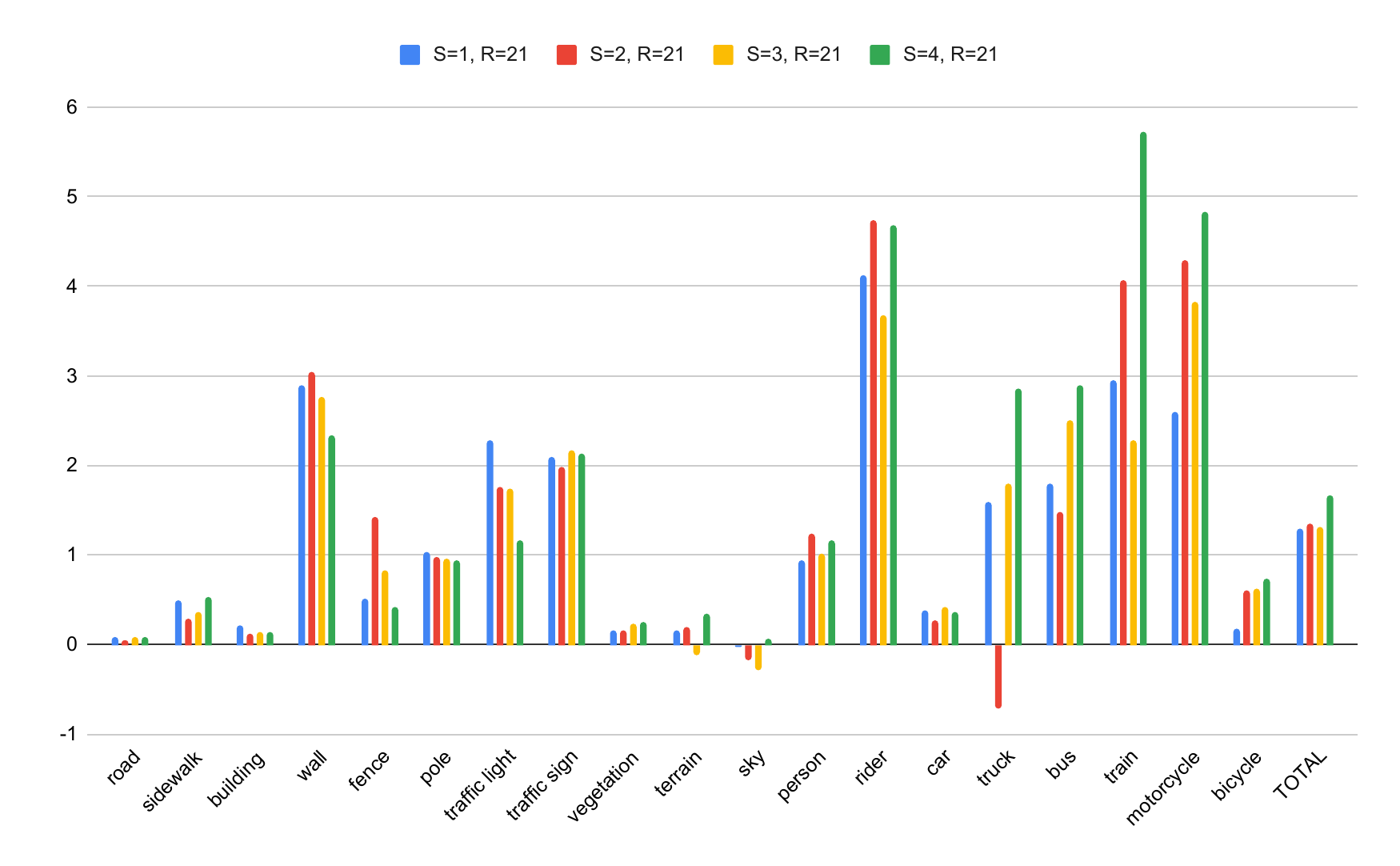}
\end{center}
\caption{Per class IoU difference between ERFNet and LMA (Ours) for different Memory sizes $S$ with a search radius $R=21$.}
\label{fig:classes_S}
\end{figure*}

\parsection{Influence of the Memory size $S$}
For a fixed search radius $R=21$, we analyze the impact of the Memory size on each individual class of the Cityscapes~\cite{Cordts2016Cityscapes} validation set for ERFNet~\cite{Romera2018ERFNetER}. Table~\ref{tab:suppl_classes_mem_size} summarizes the absolute IoU for each class while Figure~\ref{fig:classes_S} shows visually the relative improvements we obtain.

What stands out at first from the results is the fact that a set of classes that were already well segmented (road, building, vegetation, terrain, sky, car) do not change too much from the baseline. They mostly get improved by a small margin, or slightly degrade. Other classes on the other hand, get improvements ranging from 0.5\% to almost 6\%, which is very interesting, as many of those classes are either small and/or dynamic classes that are critical to understand the dynamics of the scene in the case of autonomous driving. Besides, it seems that the best improvements are in general yielded for bigger Memory sizes, while on average improvements are less striking.

In that category, the following classes are worth mentioning for our best operating point ($S=4, R=21$): wall (+2.34\%), traffic light/sign (+ 1.17\% / + 2.14\%), person/rider (+ 1.17\% / + 4.69\%), truck (+ 2.86\%), bus (+ 2.89\%), train (+ 5.73\%), motorcycle (+ 4.83\%). We can also note that, for that operating point, all the classes are improving compared to the single-frame baseline.

\begin{table*}[h]
\centering
\resizebox{1.0\linewidth}{!}{%
\begin{tabular}{l|c|ccccccccccccccccccc}
\toprule
Method  & Total & road & s.walk & build. & wall & fence & pole & tlight & sign & veg. & terrain & sky & person & rider & car & truck & bus & train & motorcycle & bicycle  \\  
\midrule

ERFNet  &  72.05	& 97.65	& 81.28	& 90.68	 &49.31	&54.89	&59.93	&62.31	  & 72.07 & 91.25  & 60.83  & 93.36	&75.91	 & 53.28 & 92.84  & 72.77   & 78.77	 & 63.78  & 46.35      & 71.74   \\
\midrule
LMA, $S=1$	  &  73.34	& 97.73	& 81.77	& 90.90	 &52.2	&55.41	&60.96	&64.60	  & 74.17 & 91.42  & 61.00  & 93.34	&76.85	 & 57.40 & 93.23  & 74.36   & 80.56	 & 66.74  & 48.95      & 71.92   \\
LMA, $S=2$	  &  73.41	& 97.69	& 81.57	& 90.81	 &52.36	&56.31	&60.91	&64.08	  & 74.05 & 91.42  & 61.03  & 93.19	&77.15	 & 58.01 & 93.12  & 72.06   & 80.25	 & 67.84  & 50.65      & 72.35   \\
LMA, $S=3$	  &  73.37	& 97.73	& 81.64	& 90.82	 &52.08	&55.72	&60.90	&64.06	  & 74.24 & 91.49  & 60.72  & 93.07	&76.93	 & 56.96 & 93.27  & 74.57   & 81.27	 & 66.06  & 50.17      & 72.36   \\
LMA, $S=4$	  &  73.72	& 97.73	& 81.81	& 90.83	 &51.65	&55.32	&60.87	&63.48	  & 74.21 & 91.50  & 61.17  & 93.43	&77.08	 & 57.97 & 93.21  & 75.63   & 81.66	 & 69.51  & 51.18      & 72.48   \\
\bottomrule
\end{tabular}}\vspace{1mm}
\caption{Comparison between ERFNet and LMA (Ours) for different Memory sizes $S$, with $R=21$.}\vspace{-3mm}
\label{tab:suppl_classes_mem_size}
\end{table*}

\begin{figure*}[h]
\begin{center}
\includegraphics[width=\linewidth]{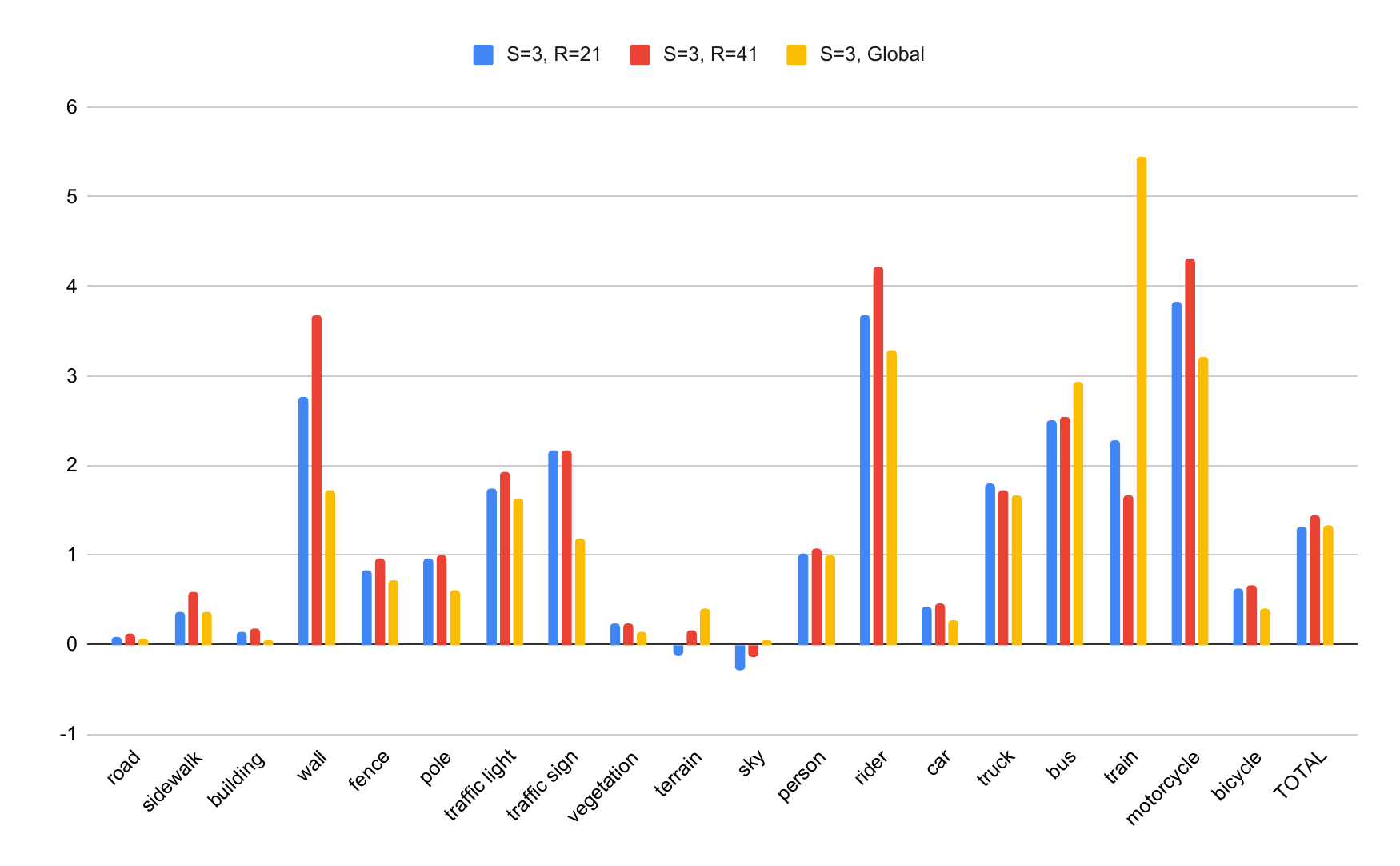}
\end{center}
\caption{Per class IoU difference between our ERFNet baseline and LMA (Ours) for different search radius $R$ with a Memory size $S=3$.}
\label{fig:classes_R}
\end{figure*}

\parsection{Influence of the Search Radius $R$}
Similarly, for a fixed Memory size $S=3$, we analyze the impact of the search radius on each individual class of the Cityscapes~\cite{Cordts2016Cityscapes} validation set for ERFNet~\cite{Romera2018ERFNetER}. Table~\ref{tab:suppl_classes_search_radius} summarizes the absolute IoU for each class while Figure~\ref{fig:classes_R} shows visually the relative improvements we obtain.

The results are consistent with the previous section, the same classes follow the same trend, for all search radii. What is most interesting to see here, is that except for the two classes \emph{train} and \emph{bus}, the all-vs-all matching yield worse results compared to a local matching, which is the behaviour we observed in our ablation study on average.

\subsection{PSPNet backbone}

\begin{figure*}[h]
\begin{center}
\includegraphics[width=\linewidth]{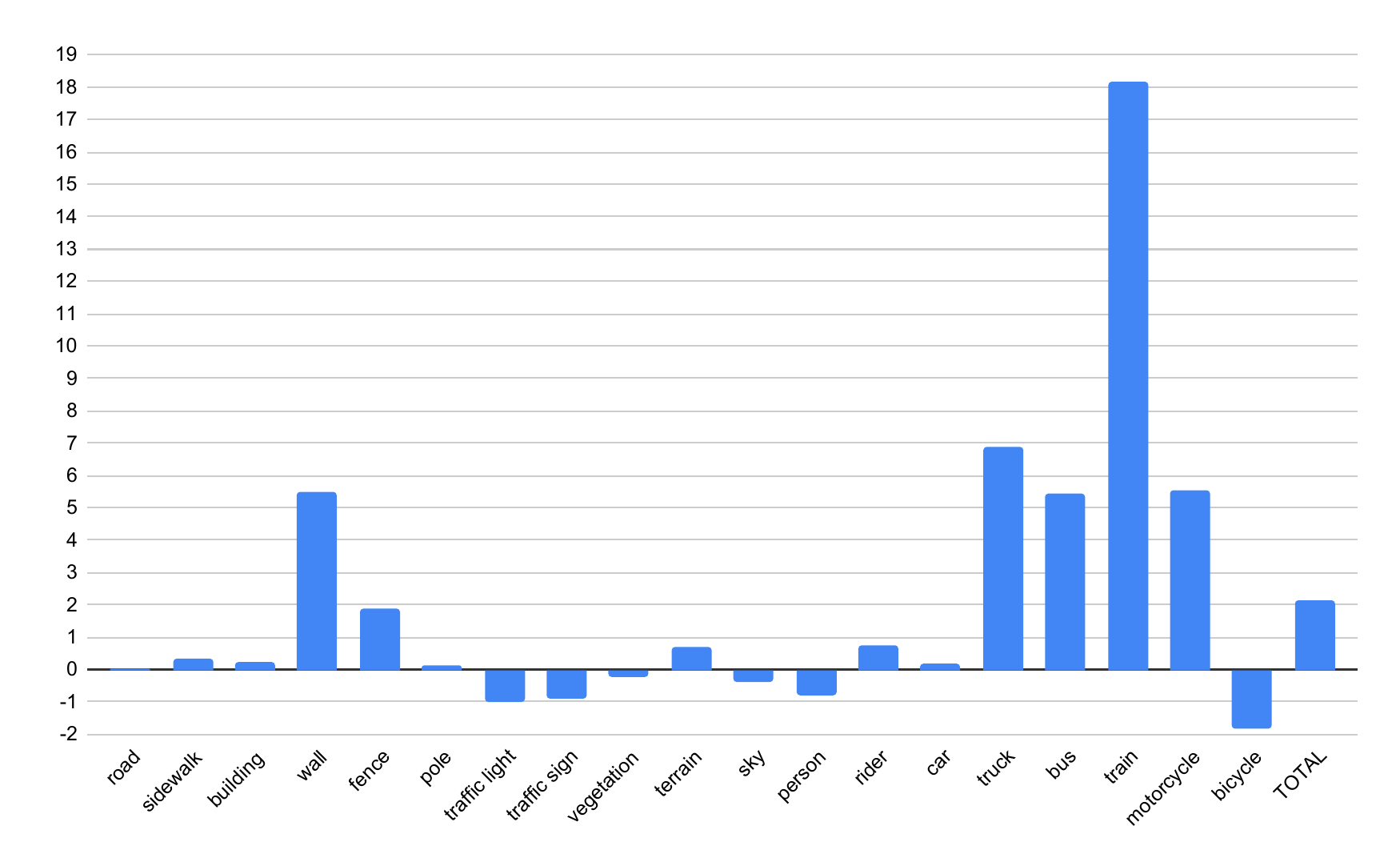}
\end{center}
\caption{Per class IoU difference between PSPNet and LMA (Ours) with $S=3$ and $R=21$.}
\label{fig:classes_psp}
\end{figure*}

Finally, we analyze the improvements obtained on the Cityscapes~\cite{Cordts2016Cityscapes} validation set for PSPNet~\cite{zhao2017pspnet}. Table~\ref{tab:suppl_classes_psp} summarizes the absolute IoU for each class while Figure~\ref{fig:classes_psp} shows visually the relative improvements we have.

In this case the trend is similar but with some noticeable differences. While bigger classes, including moving dynamic ones (train, truck, bus, motorcycle, wall) are improving substantially, smaller objects do not benefit in the same manner as before. This might be due to the stronger feature representation provided by the ResNet-101 backbone along with the higher resolution of the feature map: $256\times128\times2048$ compared to $128\times256\times128$ for ERFNet.

\begin{table*}[h]
\centering
\resizebox{1.0\linewidth}{!}{%
\begin{tabular}{l|c|ccccccccccccccccccc}
\toprule
Method  & Total & road & s.walk & build. & wall & fence & pole & tlight & sign & veg. & terrain & sky & person & rider & car & truck & bus & train & motorcycle & bicycle  \\  
\midrule

ERFNet      & 72.05	& 97.65	& 81.28	& 90.68	 &49.31	&54.89	&59.93	&62.31	  & 72.07 & 91.25  & 60.83  & 93.36	&75.91	 & 53.28 & 92.84  & 72.77   & 78.77	 & 63.78  & 46.35      & 71.74   \\
\midrule
LMA, Global & 73.38	& 97.71	& 81.65	& 90.72	 &51.04	&55.61	&60.54	&63.95	  & 73.25 & 91.39  & 61.23  & 93.41	&76.91	 & 56.56 & 93.11  & 74.43   & 81.71	 & 69.22  & 49.57      & 72.15   \\
LMA, $R=41$ & 73.50	& 97.77	& 81.86	& 90.86	 &52.99	&55.85	&60.92	&64.23	  & 74.24 & 91.49  & 61.00  & 93.22	&76.98	 & 57.50 & 93.30  & 74.49   & 81.32	 & 65.44  & 50.66      & 72.41   \\
LMA, $R=21$	& 73.37	& 97.73	& 81.64	& 90.82	 &52.08	&55.72	&60.90	&64.06	  & 74.24 & 91.49  & 60.72  & 93.07	&76.93	 & 56.96 & 93.27  & 74.57   & 81.27	 & 66.06  & 50.17      & 72.36   \\
\bottomrule
\end{tabular}}\vspace{1mm}
\caption{Comparison between ERFNet and LMA (Ours) for different search radius $R$, with $S=3$.}\vspace{-3mm}
\label{tab:suppl_classes_search_radius}
\end{table*}

\begin{table*}[h]
\centering
\resizebox{1.0\linewidth}{!}{%
\begin{tabular}{l|c|ccccccccccccccccccc}
\toprule
Method  & Total & road & s.walk & build. & wall & fence & pole & tlight & sign & veg. & terrain & sky & person & rider & car & truck & bus & train & motorcycle & bicycle  \\  
\midrule

PSPNet      &  76.34 &  98.05 & 84.71 &	92.47 &	54.66 &	60.71 &	64.05 &	71.22 &	79.09 &	92.64 &	63.49 &	94.71 &	82.46 &	62.43 &	94.96 &	73.87 &	82.51 &	58.70 &	61.70 &	78.04	\\
\midrule
LMA, $S=3$	&  78.48 &  98.06 &	85.06 &	92.73 &	60.13 &	62.57 &	64.20 &	70.22 &	78.20 &	92.40 &	64.17 &	94.35 &	81.68 &	63.16 &	95.14 &	80.77 &	87.93 &	76.86 &	67.24 &	76.21   \\
\bottomrule
\end{tabular}}\vspace{1mm}
\caption{Comparison between PSPNet and LMA (Ours) with $S=3$ and $R=21$.}\vspace{-3mm}
\label{tab:suppl_classes_psp}
\end{table*}

\section{Influence of the Memory size during inference}
\label{part3-robustness}

This last experiment consist in evaluating the impact of performing inference with a different Memory size than the model was trained for. In this example, we trained our model with $S = 4$, and try the inference with bigger Memory sizes, as shown in Table~\ref{tab:suppl_inference_mem_size}.

This shows the robustness of our approach, since the accuracy does not degrade with an increased Memory size. In fact, we even see a slight improvement when using a much larger Memory, meaning that our method can still make use of the additional information in a meaningful way.

\begin{table*}[h]
\centering
\resizebox{1.0\linewidth}{!}{%
\begin{tabular}{lcccccccccccccccccc}
\toprule
Method          & $S=4$ & $S=5$ & $S=6$ & $S=7$ & $S=8$ & $S=9$ & $S=10$ & $S=11$ & $S=12$ & $S=13$ & $S=14$ & $S=15$ & $S=16$ & $S=17$ & $S=18$ & $S=19$ \\ \midrule
LMA, $R=21$        & 73.72 & 73.69 & 73.73 & 73.72 & 73.70 & 73.72 & 73.71  & 73.72  & 73.72  & 73.73  & 73.75  & 73.77  & 73.78  & 73.70  & 73.78  & 73.77  \\ \bottomrule
\end{tabular}}\vspace{1mm}
\caption{Impact of the size of the Memory during inference on the final mIoU(\%), based on a model trained with an ERFNet backbone and a Memory size of $S=4$. }
\vspace{-3mm}
\label{tab:suppl_inference_mem_size}
\end{table*}

\end{document}